## scientific data

**OPEN**

**DATA DESCRIPTOR**

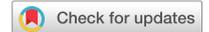

# Multimodal video and IMU kinematic dataset on daily life activities using affordable devices

Mario Martínez-Zarzuela[1]✉, Javier González-Alonso[1], Míriam Antón-Rodríguez[1], Francisco J. Díaz-Pernas[1], Henning Müller[2,3] & Cristina Simón-Martínez[2]

Human activity recognition and clinical biomechanics are challenging problems in physical telerehabilitation medicine. However, most publicly available datasets on human body movements cannot be used to study both problems in an out-of-the-lab movement acquisition setting. The objective of the VIDIMU dataset is to pave the way towards affordable patient gross motor tracking solutions for daily life activities recognition and kinematic analysis. The dataset includes 13 activities registered using a commodity camera and five inertial sensors. The video recordings were acquired in 54 subjects, of which 16 also had simultaneous recordings of inertial sensors. The novelty of dataset lies in: (i) the clinical relevance of the chosen movements, (ii) the combined utilization of affordable video and custom sensors, and (iii) the implementation of state-of-the-art tools for multimodal data processing of 3D body pose tracking and motion reconstruction in a musculoskeletal model from inertial data. The validation confirms that a minimally disturbing acquisition protocol, performed according to real-life conditions can provide a comprehensive picture of human joint angles during daily life activities.

## Background & Summary

Physical rehabilitation requires continuous monitoring to achieve a personalized exercise program that is constantly adapted to the patient's individual needs. Such monitoring has two main advantages. First, it provides the medical team with information that can be used to quantify the improvement of a specific physical therapy program. Second, it can be used to adapt online training programs to match the individual needs of the patient while reducing the needs to attend the clinic, reducing consequently the cost of healthcare. To establish such monitoring, several tele-health strategies have been proposed. Research on optical and inertial sensing devices, together with recent advances in deep learning have raised different technologies that can be used for human body tracking. The use of traditional video, acquired with a single camera, and inertial measurement units (IMUs), combined with the rapid advances in data science form the perfect scenario to profit from blending both methods to optimize tele-rehabilitation programs.

Whilst the quantitative assessment of movement analysis with patients is typically based in a laboratory with accurate high-end systems[1], the assessment of patients outside the laboratory can provide more informative measurements on their functional ability in activities of daily living[2]. Such lab-based method for quantifying human body movement consists of a multi-optoelectronic configuration of several infrared cameras (*Vicon, Qualysis, OptiTrack*), which are used for precise tracking of joint and bone markers, that are previously placed manually by an expert on the patient's body parts. However, the use of these systems in telemedicine approaches is unfeasible for several reasons, such as cost, size, configuration time, and complexity of use. It has been shown that rehabilitation in a natural environment is more effective for motor restoration compared to clinic-based programs[3]. Consequently, the use of technology in the patient's natural environment for quantitative movement tracking is crucial to meet the needs of a home-based rehabilitation program. Novel advances in computer vision and wearable devices, have shed some light onto the possibilities of performing recognition of daily life activities and kinematic evaluations in the wild using more feasible and minimally invasive solutions, that can be integrated to home-based rehabilitation approaches. In the last years, two approaches have gained attention among the scientific community: (1) single-camera systems using consumer depth cameras like *Microsoft Azure Kinect*® or 2D conventional cameras, and (2) wearable sensors using IMUs.

[1]University of Valladolid, Valladolid, Spain. [2]University of Applied Sciences and Arts Western Switzerland (HES-SO) Valais-Wallis, Sierre, Switzerland. [3]Medical faculty, University of Geneva, Geneva, Switzerland. ✉e-mail: mario.martinez@uva.es

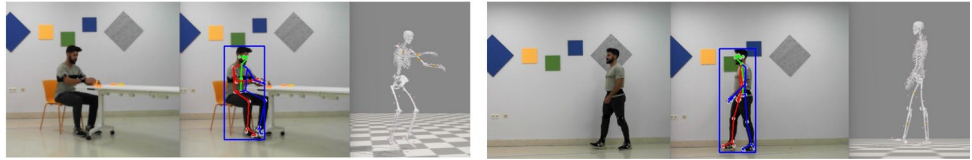

**Fig. 1** Examples of data files included in the dataset for upper (left) and lower body activities (right): a raw video file, a pose estimator from video, and a 3D motion reconstruction from inertial data. For upper-body movements (left), the subjects wear 5 IMUs in the upper limbs. For lower-body movements (right), the subjects wear 5 IMUs on the lower limbs. Individuals in the figures provided consent for their images to be published.

First, consumer depth-cameras have been widely used for patient interaction in virtual reality telerehabilitation systems[4] showing satisfactory performance in monitoring basic human poses. In addition, recent human pose estimators using deep neural networks can infer a simplified skeletal model of the human body even from 2D videos, despite the presence of cluttered backgrounds. The evolution of computer vision techniques already available in *OpenPose*[5] or *DeepLabCut*[6], among others, will eventually revolutionize the assessment of patients in their natural environment. Remarkably, they have been successfully used in neurological disorders to estimate gait parameters[7–9] and describe trunk deficits[10].

Second, inertial sensors also offer a promising alternative to gold standard movement acquisition tools. Some commercial systems, such as *Xsens Awinda* have been shown to provide joint angle measurements with a technological error under 5° RMSE with respect to multi-optoelectronic systems[11,12], yet considered today the gold standard. The use of this kind of sensors in the medical field has been thoroughly explored in an extensive number of scientific publications[13,14] proving their usability and user acceptance. Additionally, advanced signal processing tools have enabled the instrumentation of standard clinical tests that provide relevant data on balance[15] and risk of falls[13,16].

Although single-camera and inertial sensing solutions are promising, they are not without limitations. Human pose estimators using only one camera do not capture subtle movements and their joint positions detection are not yet accurate enough for 3D kinematics, due to the inherent limitations of 2D video analysis and self-occlusions of body parts[5,6]. On the other hand, although the use of inertial sensors is beneficial for tracking the 3D rotation of each body segment with high accuracy, IMU-to-segment calibration and drift challenges introduce complexity and errors in the acquisition[12].

A search among publicly available datasets on human body movement showed us that there are databases with a human activity recognition focus and those with a pure biomechanical focus. Significantly, the first ones do not use tools to reconstruct movement tracking[17–20]; and the second ones use high-end laboratory equipment, thus do not include data that could be feasibly collected in the natural environment[21–25]. Therefore, there is a need for datasets that include movements resembling daily-life activities with technologies that can be used in the wild to pave the way toward more efficient telemedicine settings that are able to recognize the patients' activity and track their movements in their natural environment.

Here, we propose the VIDIMU dataset[26] that includes 54 healthy young adults recorded with video and 16 of them simultaneously with IMUs while performing daily life activities. The novelty of the dataset is threefold: (i) the clinical relevance of the chosen movements, as these are included in typical functional assessment scales and physical rehabilitation programs, (ii) the acquisition of multimodal data has been done using very affordable equipment: a commodity webcam and custom IMU sensors, (iii) the use of open-source state-of-the-art tools for processing and synchronization of raw data.

Altogether, our dataset aims to achieve 3D body pose tracking from video and 3D motion reconstruction in musculoskeletal models from inertial data during daily-life movements and it is anticipated to contribute to advancements in various scientific domains, including human body tracking, movement forecasting and recognition, and gross motor movement assessment, among others. This valuable resource has the potential to drive the development of affordable and dependable solutions for patient monitoring in their natural environments.

## Methods

**Overview of VIDIMU.** The VIDIMU dataset[26] includes 54 healthy young adults that were recorded on video. A subgroup of 16 subjects were simultaneously recorded using IMUs. For each subject, 13 activities were registered using a low-resolution video camera and five Inertial Measurement Units (IMUs). Inertial sensors were placed in the lower or the upper limbs of the subject, respectively for activities that involve movement with the lower or the upper body. Video recordings were postprocessed using the state-of-the-art pose estimator *BodyTrack* (included in Maxine-AR-SDK[27]) to provide a sequence joint positions for each movement. This estimator was chosen for our dataset instead of other approaches such as *OpenPose*[5], because it can infer the 3D position $(x,y,z)$ of the joints from a single camera video. Raw IMU recordings were post-processed to compute joint angles by inverse kinematics with *OpenSim*[28] (see Fig. 1). In addition, for recordings including simultaneous acquisition of video and IMU data types, these signals were used for data file synchronization.

**Subjects and ethical requirements.** The data were recorded from 54 healthy adult subjects recruited among students (36 males, 18 females; 46 right-handed, 8 left-handed; age 25.0 ± 5.4 years). Before data acquisition, each subject received both written and oral explanation of the experiment and signed an informed consent form allowing their video and IMU data records to be published. The study was approved by the Institutional Review Board (or Ethics Committee) of "CEIm ÁREA DE SALUD VALLADOLID ESTE" (Valladolid, Spain), under protocol code PI 21-2341. The ethics approval allowed for the data (both video and IMU data records) to be published under an open license. Dataset acquisition took place in the facilities of the Higher School of Telecommunications Engineering of the University of Valladolid from June 2022 to January 2023. Subjects not wearing IMUs did wear a face mask because their data collection was done right after the COVID-19 pandemic. Subjects wearing IMUs were captured later and do not wear a mask.

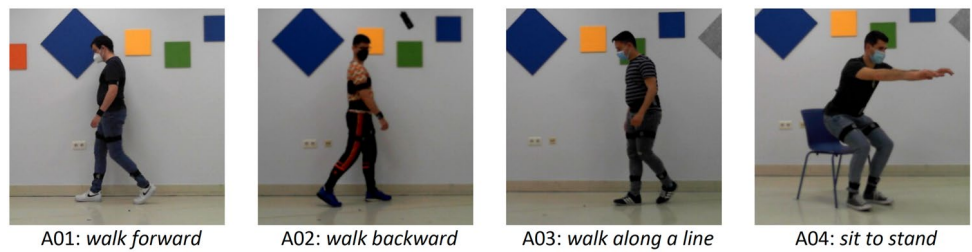

**Fig. 2** Lower limb activities in the VIDIMU dataset[26]. Individuals in the figures provided consent for their images to be published.

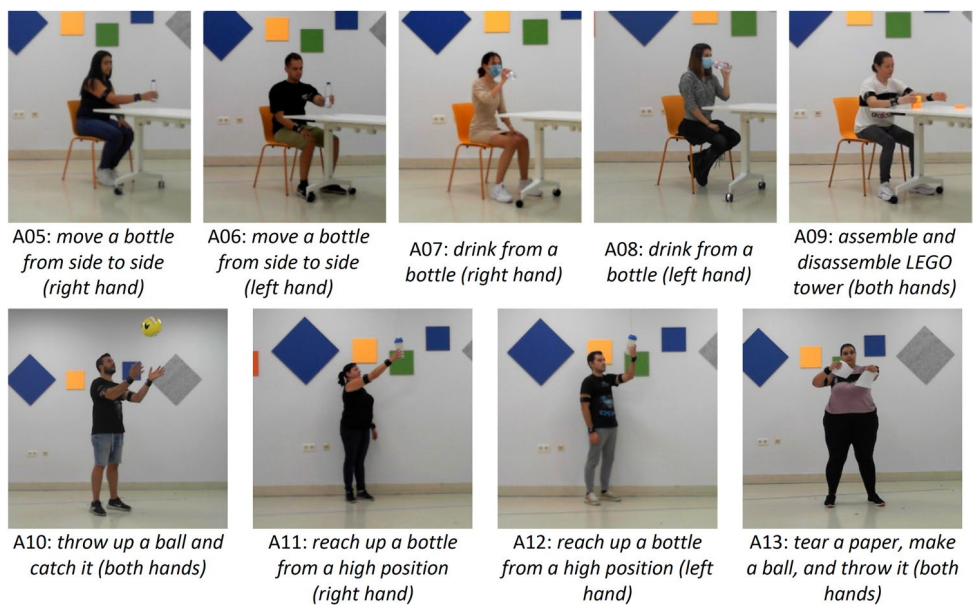

**Fig. 3** Upper limb activities in the VIDIMU dataset[26]. Individuals in the figures provided consent for their images to be published.

A battery of lower (Fig. 2) and upper (Fig. 3) limb activities, typically used to evaluate motor deficits and success of rehabilitation programs[29–32] were selected. Among the lower limb activities, we included '*walk forward* (A01)', '*walk backward* (A02)', '*walk along a line* (A03)' and '*sit to stand* (A04)'. Walking in different directions is one of the main goals of many rehabilitation programs as it forms the basis of mobility and has several positive physiological effects. The activity 'sit-to-stand' intended to mimic the transfer from being seated to a standing position, and it is also key for mobility and to gain strength in the lower limbs.

Among the upper limb activities, we included unimanual and bimanual tasks, to cover the different aspects of an upper limb rehabilitation program. As the upper limb involves movements from very simple to very complex, we attempted to cover a variety of complexities, starting from simply '*move a bottle from side to side* (*right* (A05) and *left* (A06) hand)', continuing to functional movements like '*drink from a bottle* (right (A07) and left (A08) hand)' and '*reach up a bottle from a high position* (mimicking a shelf, right (A11) and left (A12) hand). More complex and bimanual movements included '*assemble and disassemble a 6-pieces LEGO tower* (A09)', '*throw up a ball and catch it* (A10)' and '*tear a paper in 4 pieces, make a ball and throw it* (A13)'. These more complex activities are often used in rehabilitation programs to increase the leisure component of the exercises and motivate the patients.

**Acquisition setup.** The acquisition of the movements was performed using a commodity webcam (Microsoft™ LifeCam studio for business) and 5 affordable custom designed IMU sensors[33]. Video was captured

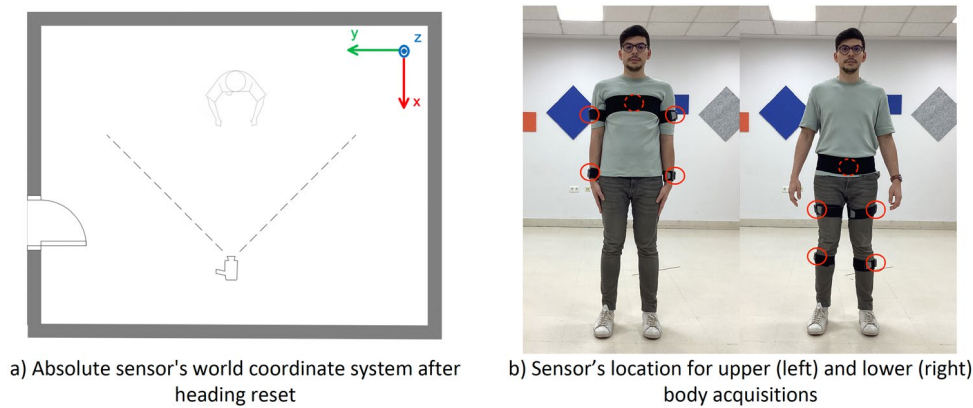

a) Absolute sensor's world coordinate system after heading reset

b) Sensor's location for upper (left) and lower (right) body acquisitions

**Fig. 4** IMU sensor's reference system and sensors placement. Individual in the figure provided consent for his image to be published.

at 30 fps and 640 × 480 pixel resolution. IMU data was acquired wirelessly at 50 Hz using the 2.4 GHz frequency band. The sensors collected quaternion data and were set according to a right-handed ENU (East North Up) coordinate system. Before the sensors were worn by the subject, they were arranged in parallel on a table and a heading reset was performed, so that the local X axis of the sensor faced towards the camera frontal plane. Figure 4a) shows the IMU reference coordinate system used during IMUs acquisition and Fig. 4b) shows sensors location for upper and lower body acquisitions.

The IMUs were placed on the subject's limbs and trunk with velcro straps. For upper body activities, the sensors were positioned on the back following an imaginary line connecting both posterior axillary folds (around T5-T7), on the lateral middle part of each upper arm, and posterior part of each wrist (Fig. 4b, left side picture). For lower body activities, the sensors were positioned on the lower back (around L3-L5), lateral middle part of each thigh, and lateral cranial part of each lower leg (Fig. 4b, right side picture). For data records including IMU data, the subject adopted a neutral pose (N-pose) before starting the movement and the instantaneous orientations of the sensors were recorded. The orientation of the sensors in this position is registered in the dataset as frame 0 and is used to perform IMU to segment calibration. A detailed description on the mathematical procedure followed can be found in a previous study[33]. In addition, the VIDIMU dataset[26] includes a video file of the subject while adopting the N-pose for each activity, and a file with the estimated joint locations during this pose.

**Acquisition protocol.** The VIDIMU dataset[26] includes data files of the 13 activities shown in Figs. 2, 3. Table 1 summarizes complementary information regarding the activities conducted by the subjects, including the number of repetitions and the oral instructions given to the participant. The instructions for the subjects to adopt N-pose were: *"adopt a standing position, facing frontally towards the camera, with the arms outstretched along the body, and with the palms of the hands facing inwards"*. It is important to note that in activities A01 and A02, the subject was moving perpendicular to the camera plane, so that gait movements were captured from the sagittal plane. In activity A03 the subjects walked along a line oriented 20 degrees with respect to the camera plane. In activities A04 to A13, the subject was turned to their left approximately 45° with respect to the frontal camera plane. This configuration has significant implications for the accuracy of the body pose detector from video. More specifically, although it avoids self-occlusions of the body, a side-effect on the measurement of joint angles is introduced.

**Signal processing.** In VIDIMU dataset[26] it is provided both the raw data and the pre-processed data. The pre-processing steps include estimating joints positions from video data and estimating joint angles from IMU data. For those subjects captured both with video and IMUs, the outputs of these processing steps were also employed to further reprocess and synchronize IMU and video files. A detailed description of these procedures follows.

The raw video data captured with the webcam was used for estimating 3D (x, y, z) body joint absolute positions in mm from video data, using *BodyTrack* from *Maxine-AR-SDK*[27]. The plain text output of *BodyTrack* was redirected to a text file (.out). The dataset[26] includes the same information in a more convenient comma-separated-values files (.csv).

The raw quaternion data captured with the IMUs (.raw) was used for inverse kinematic computation of the joint angles using *OpenSim*[28]. To this end, an *OpenSim* full-body model from Rajagopal *et al.* model[34] was edited. The model was modified to adopt a neutral pose and the constraints of the different joints were set according to Table 2.

*OpenSim's* IMU placer tool was used to orient the IMUs on the model according to the initial orientation of the real sensors during N-pose calibration. The rest of the data records in each activity were used to compute inverse kinematics (IK). The weights for IK processing were configured down-weighting distal IMUs, what improves accuracy of the kinematic estimates and reduces drift[35]. Table 3 subsumes the ideal orientation of the sensors during N-pose calibration, and the weight employed during IK.

**Data synchronization.** For those subjects and activities in which video and IMU data records were acquired simultaneously, a step for data synchronization was applied.

| ID | Activity | Reps. | Oral instructions for the subject |
|---|---|---|---|
| A01 | walk forward | 3 | "Stand on the mark placed on the ground, turn to your right and walk back and forth in a straight line between the two marks. Make the turns looking at the camera and at a slower pace" |
| A02 | walk backward | 3 | "Stand on the mark placed on the ground, turn to your left and walk back and forth in a straight line between the two floor marks. Make the turns looking at the camera and at a slower pace. You can turn your head to check if you are reaching the marks" |
| A03 | walk along a line | 3 | "Stand on the mark placed on the ground, turn to your right and walk back and forth on the line. Walk putting one foot in front of the other as if you were balancing, but without putting them together. Make the turns looking at the camera and at a slower pace" |
| A04 | sit to stand | 5 | "Starting from a sitting position, get up and sit again, without using your arms for support" |
| A05, A06 | move a bottle from side to side | 5 | "Start from a sitting position, with your hands on your knees. Reach out your right/left hand to grab the bottle that sits on the table and take it to the other mark situated on it. Then put the bottle back in its original position. Finally, return to your starting position, with your hands on your knees". |
| A07, A08 | drink from a bottle | 5 | "Start from a sitting position, with your hands on your knees. Reach out your right/left hand to grab the bottle that sits on the table and make the gesture of drinking by taking it close to your mouth, but without contact. Then bring it back to its original position on the table. Finally, return to your starting position, with your hands on your knees" |
| A09 | assemble and disassemble LEGO tower | 1 | "Start from a sitting position, with your hands on your knees. Build a tower fitting 6 pieces, 3 are placed on your right and 3 on your left. Start building with your dominant hand and alternate hands. Once the construction is finished, proceed to disassemble the tower: start with your dominant hand and alternate hands placing the pieces back on your right and left. Finally, return to your starting position, with your hands on your knees" |
| A10 | throw a ball up and catch it. | 10 | "Stand on the mark placed on the ground with your arms down, holding the ball with both hands. Throw the ball up, catch it, and return to the starting position" |
| A11, A12 | reach up a bottle from a high position | 5 | "Stand on the mark placed on the ground, with your arms down. Stretch your right/left arm to grasp the bottle situated on an elevated position. Once you have grabbed the bottle, release it and return to the starting position" |
| A13 | tear a paper, make a ball and throw it | 1 | "Stand on the mark placed on the ground with your arms down, holding the paper with both hands. Raise the paper until your arms form a 90° angle with your trunk. With both hands, break the paper into 4 pieces. Then make a ball with them and throw it" |

**Table 1.** From left to right, the information displayed includes the activity ID, activity description, number of repetitions requested from participants, and oral instructions given to the subjects before initiating the activity.

| Model joints | Min. val. | Max. val. | Model joints | Min. val. | Max. val. |
|---|---|---|---|---|---|
| pelvis_tilt | −90 | 90 | lumbar_extension | −90 | 90 |
| pelvis_list | −90 | 90 | lumbar_bending | −90 | 90 |
| pelvis_rotation | unclamped | | lumbar_rotation | −90 | 90 |
| hip_flexion | −30 | 120 | arm_flex | −90 | 180 |
| hip_adduction | −50 | 30 | arm_add | −180 | 90 |
| hip_rotation | −60 | 40 | arm_rot | −90 | 100 |
| knee_angle | −10 | 120 | elbow_flex | −10 | 180 |
| ankle_angle | blocked | | pro_sup | −10 | 180 |
| subtalar_angle | blocked | | wrist_flex | blocked | |
| mtp_angle | blocked | | wrist_dev | blocked | |

**Table 2.** Joint angles constraints in *OpenSim* model.

Firstly, for each trial the joint positions extracted from video were employed to compute the angle of a joint of interest (see Table 4). This angle was computed as the angle between two consecutive 3D body segments, which were determined from the *BodyTrack's* estimated position of the joints included in the third column of the table. The criteria to choose the joint of interest was to select the one for which the estimation of body segments could be, beforehand, more reliable according to the position of the subject and the direction of movement with respect to the camera. Joint angles for flexion-extension of the knee, the elbow, and the arm were chosen.

Secondly, joint angle signals computed from video and the IMUs were synchronized for each data record. This process included steps for: subsampling IMUs joint angles from 50 Hz to 30 Hz, using a moving average filter of 5 samples to smooth the video signals, and shifting one signal over the other until minimizing the RMSE for the first 180 signal samples, what corresponds to the 6 first seconds of the activity. A detailed view of each step is visualized in Fig. 5.

Following this strategy, synchronized versions of text files containing video (.csv) and IMU (.raw,.mot) information were generated. The VIDIMU dataset[26] includes those files in a specific subfolder (/dataset/videoandimusync). In addition, equivalent plots as those in Fig. 5, but for every subject and activity are also available as dataset files in a specific subfolder (/analysis/videoandimusync).

## Data Records

The VIDIMU dataset[26] is stored in Zenodo. Human body movements dataset record files are named according to the pattern: S##_A&&_T$$.@@@, for subject (S), activity (A) and trial (T) and where ## digits refer to the subject number, && refer to the activity, $$ refer to the recorded trial, and @@@ refer to the file extension (e.g. S40_A01_T01.raw, S40_A01_T01.mp4, S40_A01_T01.csv). Only one trial per subject and activity is available in the dataset. This single file includes all the movement repetitions required according to the protocol (see Table 1). The reason a trial identifier can be different from T01 is that some trials were discarded during acquisition due to e.g., incorrect calibration, incorrect movements, sensor, or video errors. The following body measurements of the 16 subjects recorded in video and wearing IMU sensors are provided in the "bodyMeasurements.csv" file: height (cm), weight (cm), shoulder height (cm), shoulder width (cm), elbow span (cm), wrist span (cm), arm span (cm), hip height (cm), hip width (cm), knee height (cm), ankle height (cm), foot length (cm). For the inverse kinematics process to be executed, *OpenSim* requires that the quaternion information in .raw files is translated to storage.sto file with a tabular text format, which are also included as part of the dataset[26]. The inverse kinematics process in *OpenSim* generates for every subject and record a motion.mot file that includes the prefix 'ik_' (e.g. ik_S40_A01_T01.mot), and a orientations errors .sto file that includes the same prefix and the suffix '_orientationErrors' (e.g. ik_S40_A01_T01_orientationErrors.sto).

| Lower body activities | | | Upper body activities | | |
|---|---|---|---|---|---|
| Model IMUs | Ideal orientation (w,x,y,z) | IK weigh | Model IMUs | Ideal orientation (w,x,y,z) | IK weigh |
| *pelvis_imu* | (0.7071, 0.0000, −0.7071, 0.0000) | 1.00 | *torso_imu* | (0.7071, 0.0000, −0.7071, 0.0000) | 1.00 |
| *femur_r_imu* | (0.5000, 0.5000, −0.5000, 0.5000) | 1.00 | *humerus_r_imu* | (0.5000, 0.5000, −0.5000, 0.5000) | 1.00 |
| *tibia_r_imu* | (0.5000, 0.5000, −0.5000, 0.5000) | 0.25 | *radius_r_imu* | (0.5000, 0.5000, −0.5000, 0.5000) | 0.25 |
| *femur_l_imu* | (0.5000, −0.5000, −0.5000, −0.5000) | 1.00 | *humerus_l_imu* | (0.5000, −0.5000, −0.5000, −0.5000) | 1.00 |
| *tibia_l_imu* | (0.5000, −0.5000, −0.5000, −0.5000) | 0.25 | *radius_l_imu* | (0.5000, −0.5000, −0.5000, −0.5000) | 0.25 |

**Table 3.** Ideal expected orientation of IMU sensors during N-pose calibration and weight assigned to the sensor during IK estimation.

| Activities | Joint of interest | Joint's positions | OpenSim's joint's angle |
|---|---|---|---|
| A01, A03 | left knee | *lhip, lknee, lankle* | knee_angle_l |
| A02, A04 | right knee | *rhip, rknee, rankle* | knee_angle_r |
| A05, A09 | right elbow | *rshoulder, relbow, rwrist* | elbow_flex_r |
| A06 | left elbow | *lshoulder, lelbow, lwrist* | elbow_flex_l |
| A07, A10, A11, A13 | right shoulder | *rshoulder, relbow, neck, torso* | arm_flex_r |
| A08, A12 | left shoulder | *lshoulder, lelbow, neck, torso* | arm_flex_l |

**Table 4.** Joint angles in the sagittal plane used for data synchronization. Detailed per activity, joint of interest, joint's positions (.csv files) from *BodyTrack* used to estimate the joint angle, and corresponding joint angle in *OpenSim* motion (.mot files).

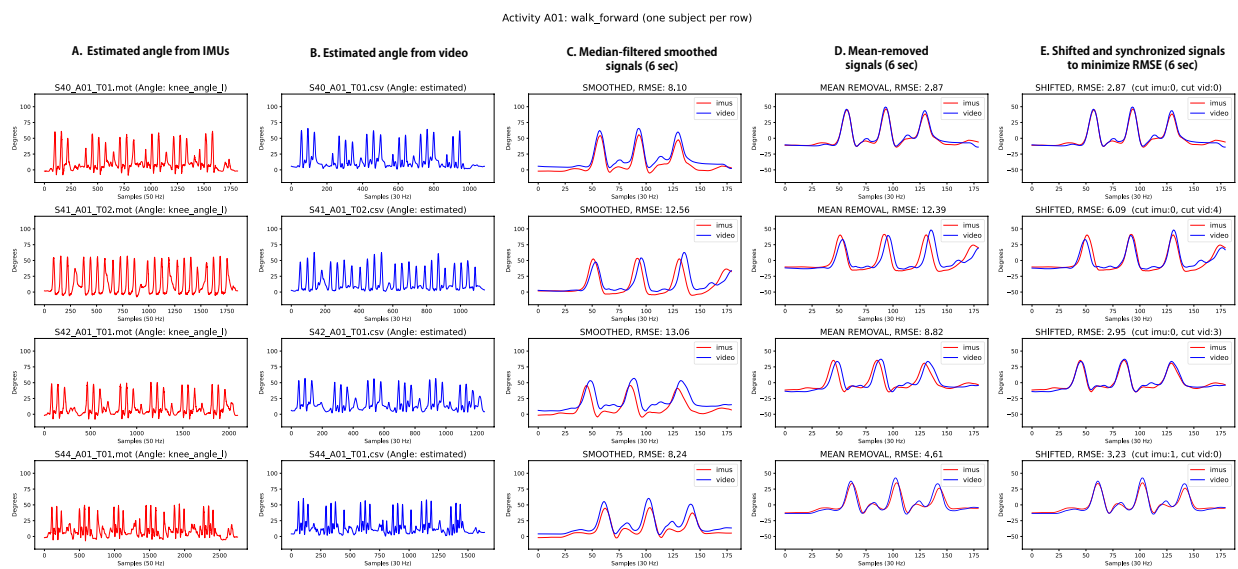

**Fig. 5** Example results of synchronization processing by minimizing RMSE between IMU (red) and video (blue) sources of data. The X axis represents the signal sample, and the Y axis represents the joint angle. For a given activity (A01 in the figure), we show the reconstructed angle with IMU data (A) and with video data (B). Panel C shows the first 180 samples (6 seconds) of the same signals after median-filtered smoothing. Panel D shows the effect of mean removal, so that the absolute ranges of motion of joint angles estimated with both sources of data can be better compared. Finally, panel E represents the optimal shifting of one of the signals, so that the Root Mean Squared Error (RMSE) is minimized. The number of samples required to shift the video or IMU signals to the left is indicated in brackets on top of the subplot in panel E: e.g. "cut imu:0, cut vid:3" would mean that the video-derived signal needs to be shifted 3 samples to the left to be synchronized with the IMU-derived signal.

| Subfolder | Size | Description |
| --- | --- | --- |
| /dataset | | Contains the file bodyMeasurements.csv including subjects anthropometric data. |
| /dataset/videoonly | <300 MB | Contains a subfolder for every subject recorded only in video:<br>-*.csv files: inferred joints positions from pose estimator (*BodyTrack*). |
| /dataset/videoandimus/ | <1GB | Contains a subfolder for every subject recorded with video and IMU sensors:<br>-*.csv files: inferred joints positions from pose estimator (*BodyTrack*).<br>-*.raw files: comma-separated-values file with originally recorded imu sensor quaternions.<br>-*.sto files: tabular storage file (*OpenSim*) containing originally recorded imu sensor quaternions.<br>-*.mot files and .sto files: tabular files containing joint angles and orientation errors generated through inverse kinematics processing (*OpenSim*). |
| /dataset/videoandimusync/ | <250MB | Contains a subfolder for every subject recorded with video and IMU sensors:<br>-*.csv files: only for those activities for which file modification was required for synchronization.<br>-*.raw files: only for those activities for which file modification was required for synchronization.<br>-*.mot files: only for those activities for which file modification was required for synchronization. |
| /analysis | <300 MB | |
| /analysis/videoonly/vangles | | For every subject recorded only in video:<br>-*.svg files: plots containing estimated joint angles computed from joint positions. |
| /analysis/videoandimus/vangles | | For every subject recorded with video and IMUs:<br>-*.svg files: plots containing estimated joint angles computed from joint positions (*BodyTrack*). |
| /analsysis/videoandimus/quats | | For every subject recorded in video and with IMUs:<br>-*.svg files: plots containing raw quaternion data. |
| /analysis/videoandimus/iangles | | For every subject recorded in video and with IMUs:<br>-*.svg files: plots containing estimated joint angles computed from inverse kinematics (*OpenSim*). |
| /analsysis/videoandimussync/ | | For every subject recorded in video and with IMUs:<br>-*.svg files: plots containing estimated synchronization requirements and final RMSE.<br>-infoToSync.csv: estimated number of frames to be cut from files for data synchronization. |
| /videosfullsize | | |
| /videosfullsize /videosoriginal | <19GB | Contains a subfolder for every subject:<br>-*.mp4 files: original video file for each activity. For subjects in the range S40 to S57 an additional video of the subject while in the N-pose is included for each activity. |
| /videosfullsize /videosbodytrack | <26GB | Contains a subfolder for every subject:<br>-*.mp4 files: output of the pose estimator (*BodyTrack*) in video.<br>-*.out files: output of the pose estimator (*BodyTrack*) to the console as plain text. |
| /videossmallsize /videosmallsize/videosoriginal | <400MB | Same as ./videosfullsize/videosoriginal, but after recoding *.mp4 files to reduce file size. |
| /videosmallsize/videosbodytrack | <800GB | Same as ./videosfullsize/videosbodytrack, but after recoding *.mp4 files to reduce file size. |

**Table 5.** Overview of dataset's folder organization.

A general overview of dataset folders hierarchy and data file formats is included in Table 5. Subjects S03, S04 revoked their informed consent for publishing their videos and data. Subjects S43 and S45 were removed because of technical issues detected during IMU data collection, and S48 was removed because of significant errors during body pose detection with *BodyTrack* caused by the poor stability of the camera focus. The related files have not been included in the dataset.

### Technical Validation

The technical validation aimed at verifying that the acquired data was representative of movement in real-life conditions and subject and activities were correctly indexed. More specifically, for video data it was checked that consistent joint angles were correctly inferred from the 3D joints using *BodyTrack*; and for IMU data it was checked that inverse kinematics generated consistent motions in a musculoskeletal model using *OpenSim*.

Detailed checking of the video data was done following several steps. First, by assuring the integrity of the files containing *BodyTrack* output. Second, by estimating joint angles and plotting them. Next, by letting a biomechanics expert to perform visual comparison of those graphs and the inferred skeleton in *BodyTrack* video output. Last, visual inspection of the coherence of plotted signals for each activity across different subjects, was used to detect possible labelling errors. Figure 6 shows an example of estimated joint angles plotting representation for subjects for lower-body activity A01 and upper-body activity A10. Joint angles estimation was performed with the code accompanying the dataset and a median filter was used to remove peaks of the signals before plotting. The dataset includes a folder with equivalent plots for every activity and subject (folders: /analysis/videoonly/vangles, and /analysis/videoandimus/vangles). It is important to note that when subject tracking errors occur, *BodyTrack* applies a default value for the 3D position of the joints. In the plots, this implies that the angle of the joints takes a constant value (e.g. 90 degrees).

The IMU data underwent comprehensive verification. First, we assured the integrity of the files containing raw IMU quaternion data. Second, we plotted the data to ensure that all the wirelessly connected sensors collected data were complete (e.g. Figure 7, more on folder: /analysis/videoandimus/quats). Next, we applied inverse kinematics using *OpenSim* and generated motion files (.mot) and plotted the estimated joint angles through inverse kinematics (e.g. Figure 8, more on folder: /analysis/videoandimus/iangles), which were verified by an expert and compared to the movements registered in video. Lastly, the generated joints graphs were visually compared with the video and IMU signals in synchronization plots (e.g. Figure 5), and the reconstructed movements were inspected in .mot files using *OpenSim* (e.g. Figure 9). Motion files (.mot) for every subject and activity are included as dataset files (folder: /dataset/videoandimus).

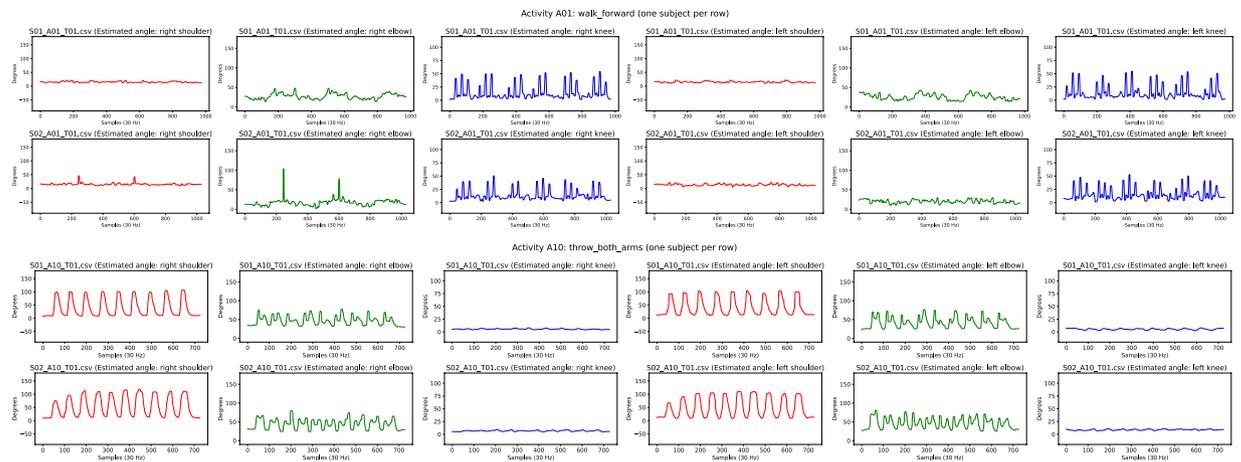

**Fig. 6** Examples of estimated joint angles inferred from 3D joint positions for activity A01 and activity A10. From left to right: *right shoulder, left shoulder, right elbow, left elbow, right knee, left knee*. Equivalent plots for every subject and activity are included as dataset files.

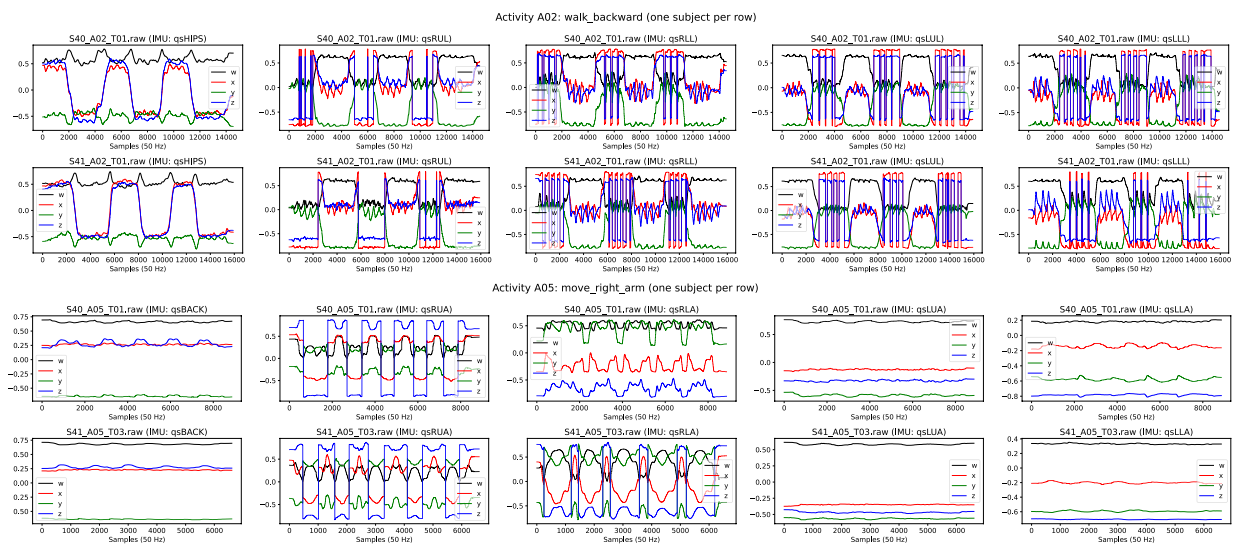

**Fig. 7** Examples of raw quaternion data collected for lower body activity A02 and upper body activity A05. For lower body activities, from left to right quaternion data from IMU sensors placed on *hips, right upper leg, right lower leg, left upper leg* and *left lower leg*. For upper body activities, from left to right quaternion data from IMU sensors placed on *back, right upper arm, right lower arm, left upper arm, left lower arm*. Equivalent plots for every subject and activity are included as dataset files.

### Usage Notes
Specific considerations on the content of the different data file types in the VIDIMU dataset[26] follow:

- .raw files: these files include original raw quaternions (w,x,y,z) info acquired with the custom IMUs. The first 5 data lines of the file contain the orientation of the IMUs while the subject was adopting the N-pose. The first column indicates the body location of the IMU sensor: *qsHIPS* stands for lower back, *qsRUL* for right upper leg, *qsRLL* for right lower leg, *qsLUL* for left upper leg, *qsLLL* for left lower leg, *qsBACK* for upper back, *qsRUA* for right upper arm, *qsRLA* for right lower arm, *qsLUA* for left upper arm, and *qsLLA* for left lower arm.
- .sto files: these files follow the required tabular format in *OpenSim* to store time series data. The dataset includes .sto files containing the same information as .raw files, and also .sto files generated by *Opensim* after the inverse kinematics computation and containing orientation errors.

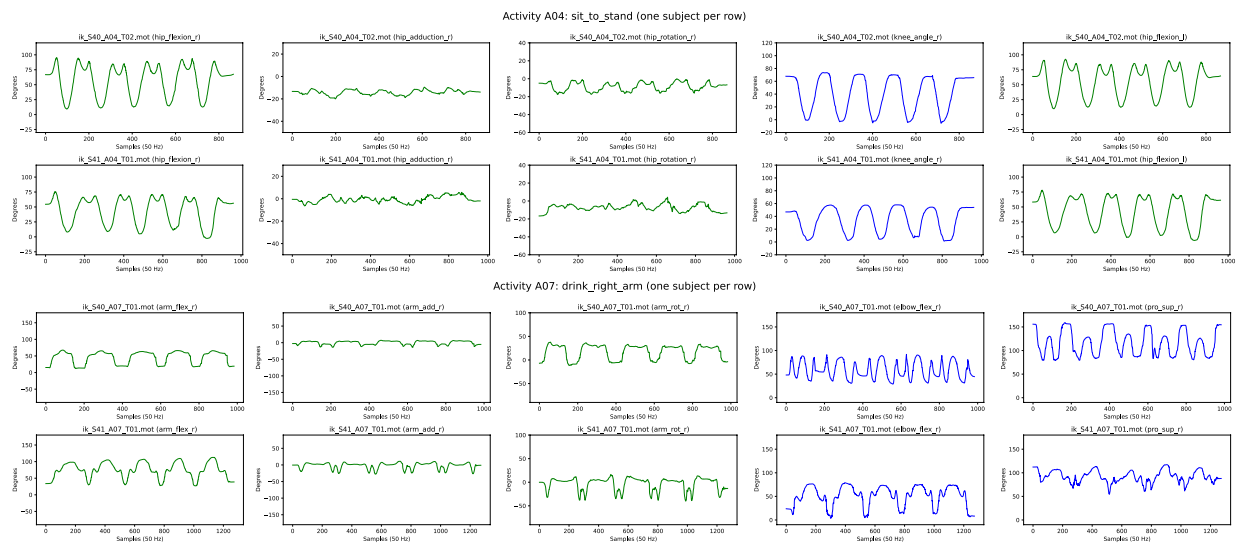

**Fig. 8** Examples of estimated joint angles computed through inverse kinematics from raw IMU data, for lower body activity A04 and upper body activity A07. Equivalent plots for every subject and activity including additional joint angles are included as dataset files. For lower body activities, these files include joint angles for: *pelvis_tilt, pelvis_list, pelvis_rotation, hip_flexion_r, hip_adduction_r, hip_rotation_r, knee_angle_r, hip_flexion_l, hip_adduction_l, hip_rotation_l, knee_angle_l*. For upperbody activities, they include joint angles for: *lumbar_extension, lumbar_bending, lumbar_rotation, arm_flex_r, arm_add_r, arm_rot_r, elbow_flex_r, pro_sup_r, arm_flex_l, elbow_flex_l, pro_sup_l*.

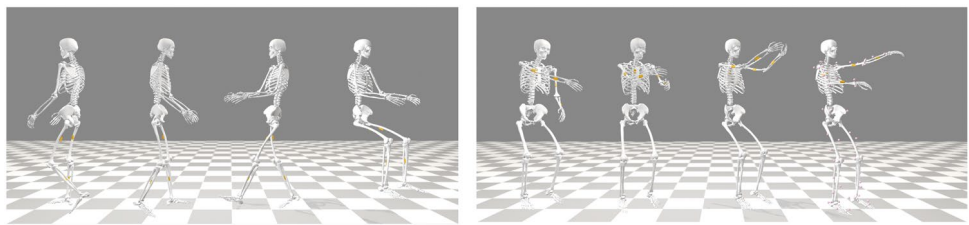

**Fig. 9** Reconstruction of movements using inverse kinematics in *OpenSim* for subject S40. Left: lower body activities A01, A02, A03, and A04. Right: upper body activities A05, A09, A10, A13. Motion files (.mot) for every subject and activity are included as dataset files.

- .mot files: these files include 3D joint angles computed in *OpenSim*. The first data line of the text file do contain the orientation of the IMUs while the subject was adopting the N-pose. These files include the following estimated joint angles in *OpenSim*: *pelvis_tilt, pelvis_list, pelvis_rotation, pelvis_tx, pelvis_ty, pelvis_tz, hip_flexion_r, hip_adduction_r, hip_rotation_r, knee_angle_r, knee_angle_r_beta, ankle_angle_r, subtalar_angle_r, mtp_angle_r, hip_flexion_l, hip_adduction_l, hip_rotation_l, knee_angle_l, knee_angle_l_beta, ankle_angle_l, subtalar_angle_l, mtp_angle_l, lumbar_extension, lumbar_bending, lumbar_rotation, arm_flex_r, arm_add_r, arm_rot_r, elbow_flex_r, pro_sup_r, wrist_flex_r, wrist_dev_r, arm_flex_l, arm_add_l, arm_rot_l, elbow_flex_l, pro_sup_l, wrist_flex_l, wrist_dev_l*. Loading various motion files in *OpenSim* at once onto a previous loaded model is faster by drag & drop of the .mot files onto the Toolbar of the application.
- .csv files: include the 3D coordinates (x, y, z) in mm estimated by *BodyTrack* of the following body parts: *pelvis, left hip, right hip, torso, left knee, right knee, neck, left ankle, right ankle, left big toe, right big toe, left small toe, right small toe, right small toe, left heel, right heel, nose, left eye, right eye, left ear, right ear, left shoulder, right shoulder, left elbow, right elbow, left wrist, right wrist, left pinky knuckle, right pinky knuckle, left middle tip, right middle tip, left index knuckle, right index knuckle, left thumb tip, right thumb tip, right thumb tip*.

Given the multimodal approach of the dataset, the thorough acquisition protocol followed, and the details explained in the technical validation section, the authors consider that the VIDIMU dataset[26] is useful for a wide range of applications. However, it should also be noted that it has certain limitations that must be considered prior to its use, specifically if the purpose is to pursue clinically applicable solutions. The main limitation in this regard is related to the lack of ground truth kinematic information using a gold standard optoelectronic system. Another limitation of the dataset is that the raw IMU data only includes the quaternion information that was collected during acquisition. Although the accelerometer, gyroscope and magnetometer data were individually acquired by every single sensor, only the quaternion data (computed internally) was wirelessly sent to a computer. The reason for this was to reduce the data load over the 2.4 GHz communication, ensuring a more reliable sensor synchronization. The data fusion algorithm that is applied internally on every BNO080 sensor was configured following the manufacturer's recommendations. Among those possible configurations, dynamic data acquisition using the *rotation vector* configuration was chosen.

## Code availability

The VIDIMU dataset[26] (https://doi.org/10.5281/zenodo.7681316) was built using the free tools *BodyTrack* (v0.8) and *OpenSim* (v4.4). The VIDIMU-TOOLS code contains the Jupyter notebooks and Python scripts used for data conversion, data synchronization and checking the contents of the dataset to ensure its integrity. A first release of the VIDIMU-TOOLS project is accessible in Zenodo[36] (https://doi.org/10.5281/zenodo.7693096) and the latest version of the code can be found in GitHub (https://github.com/twyncoder/vidimu-tools).

### Acknowledgements
The authors would like to thank the volunteers who participated in the data collection. This research was partially funded by the funded by the Ministry of Science and Innovation of Spain under research grant "Rehabot: Smart assistant to complement and assess the physical rehabilitation of children with cerebral palsy in their natural environment", with code 124515OA-100, and the mobility grant "Ayudas Movilidad Estancias Senior (Salvador Madariaga 2021)" with code PRX21/00612. Cristina Simon-Martinez is funded by the European Union's Horizon 2020 research and innovation program under the Marie Sklodowska-Curie grant agreement No 890641 ("Optimizing Vision reHABilitation with virtual-reality games in paediatric amblyopia (V-HAB)").

### Author contributions
M. Martinez-Zarzuela contributed to the conceptualization of the dataset, data collection, data curation, code programming and writing of the first version of the manuscript. J. González-Alonso contributed to custom IMU sensors development, subject measurement, data collection and data processing. M. Antón-Rodríguez contributed to video data collection, video data curation and manuscript reviewing and editing. F.J. Díaz-Pernas contributed to IMU data curation and manuscript reviewing and editing. Henning Müller contributed to the conceptualization of the dataset, manuscript reviewing and editing. Cristina Simón-Martínez contributed to the conceptualization of the dataset, data curation, data validation and manuscript reviewing and editing.

### Competing interests
The authors declare no competing interests.

### Additional information
**Correspondence** and requests for materials should be addressed to M.M.-Z.

**Reprints and permissions information** is available at www.nature.com/reprints.

**Publisher's note** Springer Nature remains neutral with regard to jurisdictional claims in published maps and institutional affiliations.